\documentclass[letterpaper, 10 pt, conference]{ieeeconf}  

\IEEEoverridecommandlockouts                              

\usepackage{graphics} 
\usepackage{epsfig} 
\usepackage{amsmath}
\usepackage{amssymb}  
\usepackage{bm}  
\usepackage{makecell}
\usepackage[noadjust]{cite} 
\usepackage{url}
\usepackage{hyperref}
\usepackage{booktabs} 
\usepackage{subfigure} 
\usepackage{graphicx}
\usepackage{textcomp}
\usepackage{float}

\usepackage{lipsum}
\usepackage{algorithm}
\usepackage{algpseudocode}
\usepackage{setspace}

\usepackage{titlesec}

\titlespacing*{\section}
{0pt}{8pt plus 2pt minus 2pt}{8pt plus 2pt minus 2pt}

\algrenewcommand\algorithmicrequire{\textbf{Input:}}
\algrenewcommand\algorithmicensure{\textbf{Output:}}

\usepackage{optidef}
\usepackage{caption}
\usepackage{tikz}
\usetikzlibrary{shapes,arrows.meta,arrows,positioning,calc}

\def\ie{\emph{i.e.}} 
 
\algnewcommand{\IRequire}{\item[\textbf{Input:}]}

\DeclareMathOperator*{\argmin}{arg\,min}

\title{\LARGE \bf
A Probabilistic Motion Model for Skid-Steer Wheeled \\ Mobile Robot Navigation on Off-Road Terrains
}
\author{Ananya Trivedi,$^{1}$ Mark Zolotas,$^{1}$ Adeeb Abbas,$^{1}$ Sarvesh Prajapati,$^{1}$ Salah Bazzi,$^{1}$ and Ta\c{s}k{\i}n Pad{\i}r$^{1}$ 
\thanks{$^{1}$Institute for Experiential Robotics, Northeastern University, Boston, Massachusetts, USA. { \tt\small \{trivedi.ana, m.zolotas, abbas.a, prajapati.s, s.bazzi, t.padir\}@northeastern.edu}}
\thanks{Research was sponsored by the United States Army Corps of Engineers (USACE) Engineer Research and Development Center (ERDC) Geospatial Research Laboratory (GRL) and was accomplished under Cooperative Agreement Federal Award Identification Number (FAIN) W9132V-22-2-0001. The views and conclusions contained in this document are those of the authors and should not be interpreted as representing the official policies, either expressed or implied, of USACE EDRC GRL or the U.S. Government. The U.S. Government is authorized to reproduce and distribute reprints for Government purposes notwithstanding any copyright notation herein.}
\thanks{Python implementation of the motion model is available at: \url{https://github.com/RIVeR-Lab/multiterrain-gp-model}}
}

\begin{document}
\tikzset{
block/.style = {draw, fill=white, rectangle, minimum height=3em, minimum width=3em},
tmp/.style  = {coordinate}, 
sum/.style= {draw, fill=white, circle, node distance=1cm},
input/.style = {coordinate},
output/.style= {coordinate},
pinstyle/.style = {pin edge={to-,thin,black}
}
}
\maketitle
\thispagestyle{empty}
\pagestyle{empty}

\begin{abstract}

Skid-Steer Wheeled Mobile Robots (SSWMRs) are increasingly being used for off-road autonomy applications. When turning at high speeds, these robots tend to undergo significant skidding and slipping. In this work, using Gaussian Process Regression (GPR) and Sigma-Point Transforms, we estimate the non-linear effects of tire-terrain interaction on robot velocities in a probabilistic fashion. Using the mean estimates from GPR, we propose a data-driven dynamic motion model that is more accurate at predicting future robot poses than conventional kinematic motion models. By efficiently solving a convex optimization problem based on the history of past robot motion, the GPR augmented motion model generalizes to previously unseen terrain conditions. The output distribution from the proposed motion model can be used for local motion planning approaches, such as stochastic model predictive control, leveraging model uncertainty to make safe decisions. We validate our work on a benchmark real-world multi-terrain SSWMR dataset. Our results show that the model generalizes to three different terrains while significantly reducing errors in linear and angular motion predictions. As shown in the attached video, we perform a separate set of experiments on a physical robot to demonstrate the robustness of the proposed algorithm. 

\end{abstract}

\section{Introduction}
\label{sec:introduction}
Robots with a skid-steer drive mechanism are becoming increasingly popular for autonomous off-road navigation due to their simple mechanical structure, high traction, and large payload capabilities~\cite{SSMR_Survey}. Skid-Steer Wheeled Mobile Robots (SSWMRs) achieve turning motion by rotating one side of the wheels at different revolutions per minute than the other. As a result, they are able to turn in-place which increases their maneuverability. However, this also leads to significant tire skidding and slipping~\cite{vehicle_dynamics}. Predicting robot motion in response to commanded velocities is thus a challenging task. 

 Gaussian Process (GP) Regression~\cite{rasmussen} is a nonlinear function approximation technique that provides mean and covariance estimates for a new test point. By propagating uncertainty in the robot states forward, several high-speed safety-critical autonomy applications have used GP Regression (GPR) to design risk-aware motion planning approaches that can balance caution and aggression~\cite{ostafew,eth_gpmpc}.  

\begin{figure}
    \centering\includegraphics[width=0.95\columnwidth]
    {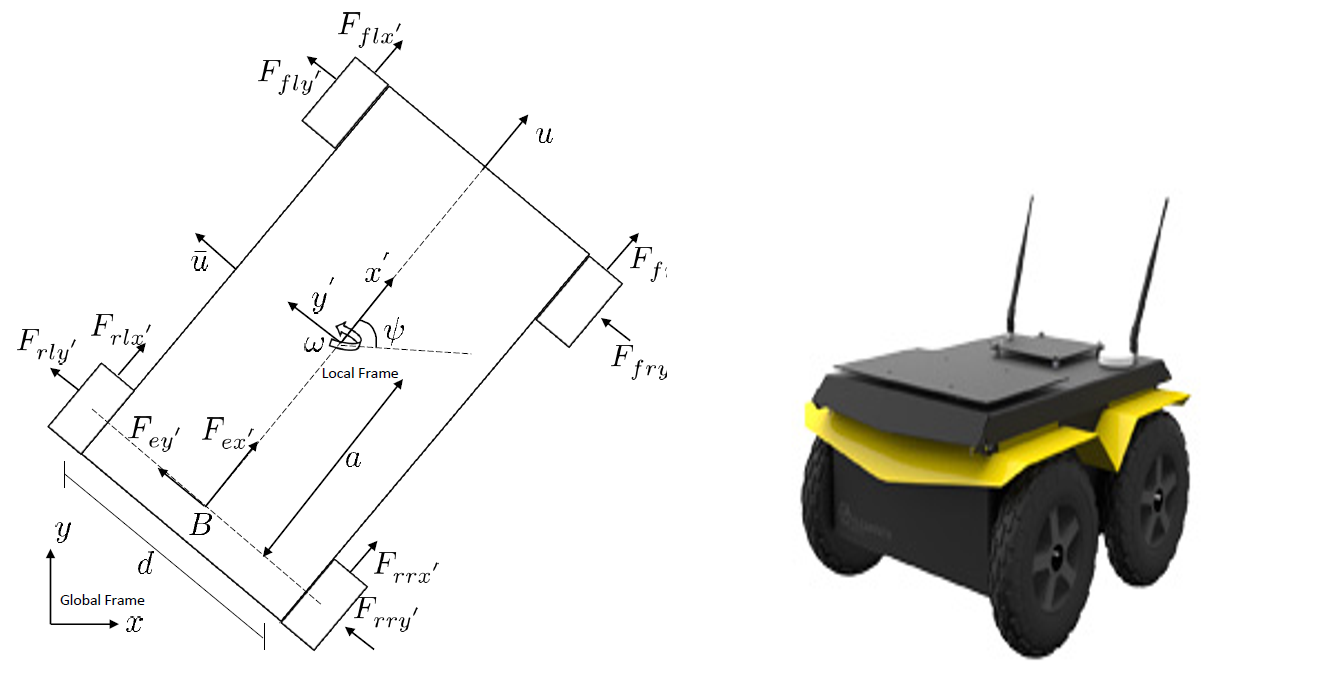}
        \caption{\textbf{Right:} Skid-Steer Wheeled Mobile Robot (SSWMR) Clearpath Jackal used in our experiments. \textbf{Left:} Free body diagram representing the forces and torques acting on an SSWMR.}
    \label{fig:jackal}
    \vspace{-0.6mm}
\end{figure}

In this work, we utilize GPR to estimate the effects of tire skid and slip for SSWMRs operating on off-road terrains, such as the one shown in Fig.~\ref{fig:jackal}. Utilizing Sigma-Point Transforms~\cite{ukf}, we approximate the distribution of robot positions and velocities in response to these effects. By computing a weighted sum of different GP outputs, we are able to generalize and adapt to unseen terrain conditions. 

In summary, our contributions are as follows:
\begin{itemize}
    \item A probabilistic motion model for SSWMRs capable of accurately estimating the distribution of robot positions and velocities in response to tire skid and slip.
    \item A convex optimization formulation to combine the GP outputs from different terrains allowing the motion model to be used for diverse, potentially unseen terrains.
    \item Experimental results on an extensive, multi-terrain SSWMR dataset demonstrating improvements in prediction performance compared to existing state-of-the-art kinematic motion models.
\end{itemize}

\section{Related Work}
\label{sec:related_work}

Tire slip is the relative motion between a tire and the road surface. For each wheel, a slip ratio $s$ and slip angle $\alpha$ can be defined as~\cite{slip_is_non_linear}:
\begin{equation} \label{eq:slip_definition}
    s = 1 - \frac{V_{wx}}{r\,\omega}, \quad \alpha = \arctan\left( \frac{V_{wy}}{V_{wx}}\right),
\end{equation}
where $V_{wx}$, $V_{wy}$ are longitudinal and lateral wheel velocities, $\omega$ is the wheel angular velocity, and $r$ is the wheel radius. Estimating $V_{wx}$, $V_{wy}$, and $\omega$ tends to be noisy in practice. In this work, we circumvent this issue by relying instead on the robot's center-of-mass (COM) position and orientation. Visual odometry methods, such as ORB-SLAM2~\cite{orb_slam}, can estimate a robot's COM pose at millimeter-level precision for different outdoor environments.

Kinematic approaches for motion modeling of SSWMRs \cite {taskin_ssmr} quantify the effects of skid and slip by relating commanded wheel velocities \(\omega_l\) and \(\omega_r\) to measured velocities \(v_x\), \(v_y\), and \(\omega\) using a  Jacobian matrix $\boldsymbol{J}$~\cite{low_and_wang_kinematic,martinez_kinematic}: 
\begin{equation}
\label{eq:kinematic_models}
    \dot{\boldsymbol{x}} = \begin{bmatrix}
        v_x \\ v_y \\ \omega
    \end{bmatrix} = 
    \boldsymbol{J}\begin{bmatrix}
        \omega_l \\ \omega_r
    \end{bmatrix}.
\end{equation}
Baril et al.~\cite{baril} recently applied five popular data-driven kinematic models to a 590\,kg SSWMR. We benchmark our proposed method against these models. The effects of tire-terrain interactions on robot motion vary with terrain, acceleration, weather, tire lifetime, etc. Nonetheless, purely kinematic models are unable to accurately capture these nuances, as we verify in Section~\ref{sec:results}. Dynamic approaches on the other hand, quantify the effects of skid and slip by explicitly modeling the forces and torques acting between the robot and the surface. However, many of these approaches are either suited for low-speed operation or are limited to validation in simulation~\cite{yu_dynamic,kozlowski_sim_dynamic,seegmiller_dynamic}. In this work we address high-speed motion modeling on varying terrain conditions.

Rabiee et al.~\cite{rabiee} developed a friction-based kinematic model that reasons about slippage and skid at the wheel level. While this motion model achieved impressive accuracy at high speeds of operation, the method presented was computationally prohibitive, as every inference step required a non-convex optimization problem to be solved. Their method further relied on measured wheel velocities, which are noisy signals. In contrast, we employ GPR for motion modeling. GPR has the advantage of being easily parallelized on a Graphics Processing Unit (GPU) using libraries, such as GPyTorch~\cite{gpytorch}. Additionally, our work generalizes to navigation over multiple terrains. 

Existing research on GPR applied to SSWMR motion planning involves customizing the unicycle kinematic model to achieve high-precision tracking of a predefined global path~\cite{ostafew,huskic,jfr_gp_sswmr}. However, these approaches are unable to perform obstacle-avoidance maneuvers and are intended for use on a single terrain. On the other hand, the general-purpose motion model introduced in this work can be integrated with navigation cost-maps. Reasoning over cost-maps, as opposed to path-following, facilitates local path planning strategies that can avoid, explore, or target particular regions in the robot's environment~\cite{teb,Everett_SSMR}.

Using an ensemble of GPs, Nagy et al.~\cite{ensemble_gp} simulated high-speed race car driving over surfaces with varying friction properties. Inspired by their approach, we extend the versatility of our motion model to previously unseen terrains. The resultant GPR estimates are propagated forward using Sigma-Point Transforms, hence providing us with a Gaussian distributed estimate of the next robot state. Metrics computed on a real-world SSWMR dataset show the efficacy of our approach.

\section{Probabilistic Motion Modeling}
\label{sec:model}

Fig.~\ref{fig:jackal} shows the forces and torques acting on an SSWMR. Throughout the text, we use bold-face to represent vectors and matrices. The \(X\), \(Y\) positions in the global frame constitute the \textit{kinematic} sub state-space, \(\boldsymbol{q} = [X, \, Y, \, \theta]^\intercal \! \in \! \mathbb{R}^3 \). The linear velocity, \(v\), and angular velocity, \(\omega\), in the local frame express the \textit{dynamic} sub state-space, \(\boldsymbol{\eta} = [v, \, \omega]^\intercal \! \in \! \mathbb{R}^2 \). The overall state-space representation is thus \(\boldsymbol{x} = [X, \, Y, \, \theta, \, v, \, \omega]^\intercal \! \in \! \mathbb{R}^5  \). The control space, \(\boldsymbol{u} = [v_{ref}, \, \omega_{ref}]^\intercal \in \mathbb{R}^2\), represents the desired linear and angular velocities.
 
\begin{figure*}[t]
    \vspace{1em}
        \centering
          \includegraphics[width=0.7\linewidth]
          {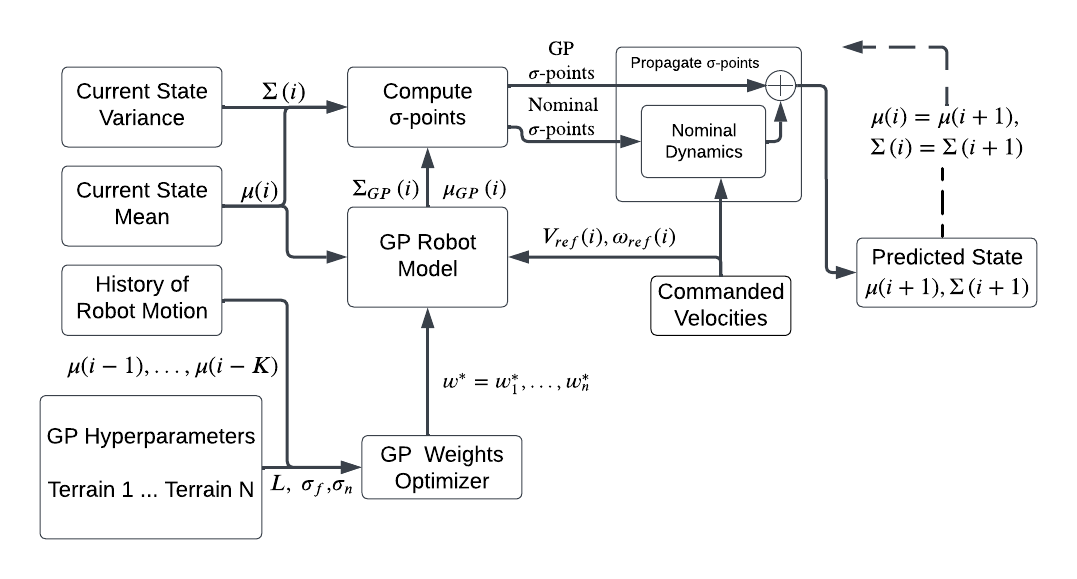}
    \captionsetup{skip=1pt} 
    \caption{Overview of our GPR approach to probabilistic motion modeling for SSWMRs navigating multiple terrains.}
    \label{fig:overview_block}
\end{figure*}

 At each time instant, $k$, the proposed probabilistic motion model takes in the current state $\boldsymbol{x}(k)$ and commanded velocities $\boldsymbol{u}(k)$ as inputs, and then predicts the distribution of the next robot state $\boldsymbol{x}(k+1)\,\sim\,\mathcal{N}(\boldsymbol{\mu}(k+1),\,\boldsymbol{\Sigma}(k+1)))$. In the remainder of this section, we outline the background and steps necessary to realize the motion model. The block-diagram of our holistic approach is outlined in Fig.~\ref{fig:overview_block}.

\setlength{\belowcaptionskip}{-8pt}
\setlength{\belowcaptionskip}{0pt}

\label{sec:method}
\subsection {Dynamic Unicycle Model}
\label{sec:unicycle}
In the following, we briefly describe the dynamic unicycle model for SSWMRs presented in Fig.~\ref{fig:jackal}. As shown in~\cite{dynamic_unicycle}, the differential equations of motion are:
\begin{equation} \label{eq:kinematic_continuous}
    \dot{\boldsymbol{q}} = 
    \begin{bmatrix}
        \dot{X}\\
        \dot{Y} \\
        \dot{\theta}
    \end{bmatrix} = 
    \begin{bmatrix}
        v \, cos\theta - a \, \omega \, sin\theta \\
        v \, sin\theta + a \, \omega \, cos\theta \\
        \omega
    \end{bmatrix},
\end{equation}
\begin{equation} \label{eq:dynamic_continuous}
    \dot{\boldsymbol{\eta}} = 
    \begin{bmatrix}
        \dot{v}\\
        \dot{\omega}
    \end{bmatrix} = 
    \begin{bmatrix}
        \frac{c_3}{c_1} \, \omega^2 - \frac{c_4}{c_1} \, v \\[4pt]
        -\frac{c_5}{c_2} \, v \, \omega - \frac{c_6}{c_2} \, \omega 
    \end{bmatrix}
    + 
    \begin{bmatrix}
        \frac{1}{c_1} \, v_{ref} \\[4pt]
        \frac{1}{c_2} \, \omega_{ref}
    \end{bmatrix}
    +
    \begin{bmatrix}
        \, \overline{\delta}_v \\[4pt] 
        \, \overline{\delta}_\omega
    \end{bmatrix}.
\end{equation}

Here, $a$ is the distance between the robot's COM and rear axle. \(c_1,\ldots,c_6\), are constants which are functions of physical robot parameters, such as motor gear ratio, moment of inertia, electric resistance, etc. These are independent of the terrain and can thus be computed once offline. As in prior work~\cite{eth_gpmpc}, we ignore the effects of lateral velocity error on the dynamic state sub-space. This error is generally small, noisy, and difficult to estimate reliably. We did not find any noticeable change in our evaluation metrics upon ignoring it. While the focus of this work is on planar robot motion modeling,  Eqn.~\ref{eq:dynamic_continuous} can be extended to account for velocities in the z-axis similar to~\cite{ssmr_modeling_3d}.

Quantities, \(\overline{\delta}_v\) and \(\overline{\delta}_\omega\) in Eqn.~\ref{eq:dynamic_continuous}, capture the cumulative effects of tire skid and slip on the velocities. In the next section, we show how these non-linear effects can be estimated using GPR. Disregarding these effects for now and rearranging Eqn.~\ref{eq:dynamic_continuous}, we obtain the following motion model, which is linear in its parameters, \(\boldsymbol{c} = [c_1,\ldots,c_6]^{\intercal}\! \in \! \mathbb{R}^6\):
\begin{equation} \label{eq:c1_to_c6}
    \begin{bmatrix}
        \dot{v} & 0 & -\omega^2 & v & 0 & 0 \\[4pt]
        0 & \dot{\omega} & 0 & 0 & v\,\omega & 0
    \end{bmatrix}
    \boldsymbol{c} = 
    \begin{bmatrix}
        v_{ref} \\[4pt]
        \omega_{ref}
    \end{bmatrix}.
\end{equation}

Applying a first order low-pass filter to both sides of Eqn.~\ref{eq:c1_to_c6} removes the dependency on noisy acceleration measurements, \(\dot{v}\) and \(\dot{\omega}\). We can then estimate \(c_1,\ldots,c_6\) by solving the resultant least squares problem~\cite{slotine,least_squares_params}.

\subsection{Estimating Effects of Tire Skid and Slip}
\label{sec:gp}
In this section, we begin by introducing the equations associated with a GPR problem. We then demonstrate how GPR can be used to estimate the effects of \(\overline{\delta}_v\) and \(\overline{\delta}_\omega\) on robot velocities.

\subsubsection{Gaussian Process Regression (GPR) Equations}
\label{sec:GPR_Basics}
We consider a vector-valued input feature vector, $\boldsymbol{z} \!\in\!\mathbb{R}^n$, and a scalar output, $y\!\in\!\mathbb{R}$.  A GPR problem can be completely specified by a kernel function, $\kappa(\cdot, \cdot)$, and its associated hyperparameters. We chose the Squared Exponential (SE) kernel function due to its property of being infinitely differentiable~\cite{car_gp_mpc}. This enables optimal control solvers, such as~\cite{acados,forcespro}, to compute gradients of robot dynamics relatively easily. The SE kernel is defined as:
\begin{equation} \label{eq:RBF}
    \kappa(\boldsymbol{z},\,\boldsymbol{z}^{\prime}) = \sigma_f^2\,\text{exp}\,\left(-\frac{1}{2}\,\boldsymbol{r}^\intercal\,\boldsymbol{L}^{-1}\,\boldsymbol{r} \right) + \sigma_n^2,
\end{equation}
where $\boldsymbol{r}=\boldsymbol{z} - \boldsymbol{z}^{\prime}$, \(\boldsymbol{L}\) is the diagonal length scale matrix, and \(\sigma_f\), \(\sigma_n\) represent the signal variance and signal noise, respectively. Given $p$ training input features, $\boldsymbol{Z}=[\boldsymbol{z}_1,\ldots,\boldsymbol{z}_p]^\intercal\!\in\!\mathbb{R}^{p \times n}$, and corresponding outputs, $\boldsymbol{y}=[y_1,\ldots,y_p]\!\in\!\mathbb{R}^{p \times 1}$, the hyperparameters, \(\boldsymbol{L}\), \(\sigma_f\), and \(\sigma_n\) can be derived using maximum likelihood estimation. 

For a new query point, \(\boldsymbol{z}_{test}\!\in\!\mathbb{R}^n\), we define: $\boldsymbol{k}_{test}=\kappa(\boldsymbol{z}_{test},\boldsymbol{z}_{test})$, and $\boldsymbol{k}\!\in\!\mathbb{R}^{1 \times p}$ as the row vector obtained by applying the kernel function between the test input and all training inputs. The matrix, $\boldsymbol{K}\!\in\!\mathbb{R}^{p \times p}$, is then obtained by applying the kernel function between all the training inputs. Therefore, the output mean and covariance estimates for $\boldsymbol{z}_{test}$ can be estimated as follows~\cite{rasmussen}:
\begin{equation}
\begin{aligned}\label{eq:gp_mean_and_covar_eq}
    \mu(\boldsymbol{z}_{test}) &= \boldsymbol{k}_{test}\!\left(\boldsymbol{K} + \boldsymbol{I}\sigma_n^2\right)^{-1}\!\boldsymbol{y}, \\
    \Sigma(\boldsymbol{z}_{test}) &= \boldsymbol{k}_{test} - \boldsymbol{k}\left(\boldsymbol{K} + \boldsymbol{I}\sigma_n^2\right)^{-1}\boldsymbol{k}^\intercal,
\end{aligned}
\end{equation}
where $\boldsymbol{I}$ is the $p \times p$ identity matrix.

\subsubsection{GPR for SSWMR Modeling}
\label{sec:GPR_SSWMR}

The discretized form of Eqns.~\ref{eq:kinematic_continuous} \&~\ref{eq:dynamic_continuous} can be expressed as:
\begin{equation}\label{eq:discretized_complete}
    \boldsymbol{x}(k+1) = 
    \boldsymbol{f}(\boldsymbol{x}(k),\boldsymbol{u}(k)) + \boldsymbol{g}(\boldsymbol{x}(k),\boldsymbol{u}(k)),
\end{equation}
where, for time instant \(k\), \(\boldsymbol{f}=[f_X,\,f_Y,\,f_\theta,\,f_v,\,f_\omega]^\intercal\) represents the nominal motion equations, while \(\boldsymbol{g}=[g_X,\,g_Y,\,g_\theta,\,g_v,\,g_\omega]^\intercal\) = \([0,\,0,\,0,\,g_v,\,g_\omega]^\intercal\) represents the disturbance motion equations due to unmodeled effects, \(\overline{\delta}_v\) and \(\overline{\delta}_\omega\). Hence:
\begin{equation}
\begin{aligned} \label{eq:model_residues}
        g_v(k) &= v(k+1) - f_v(\boldsymbol{z}(k)), \\
        g_\omega(k) &= \omega(k+1) - f_\omega(\boldsymbol{z}(k)),    
\end{aligned}
\end{equation}
with the augmented vector, \(\boldsymbol{z}(k)\), defined as \([v(k),\,\omega(k),\,v_{ref}(k),\,\omega_{ref}(k)]^\intercal\), and \(v(k+1)\), \(\omega(k+1)\) as the next state ground velocities. 

Using the mean and covariance estimates from Eqn.~\ref{eq:gp_mean_and_covar_eq}, we can train two independent GPR models with input vector, \(\boldsymbol{z}(k)\), and respective output vectors, as defined in Eqn.~\ref{eq:model_residues}. 
Given robot velocities and commands, \(\boldsymbol{\bar{z}} = [\bar{v}(k),\,\bar{\omega}(k),\,\bar{v}_{ref}(k),\,\bar{\omega}_{ref}(k)]^\intercal\) yields a Gaussian distributed approximation of the effects of tire skid and slip from the two GPs at time-step $k$ as:
\begin{equation}
\begin{aligned}\label{eq:GP_approx}
    g_v(k) = g_v(\boldsymbol{\bar{z}}(k)) &\sim \mathcal{N}(\mu_v(k),\Sigma_v(k)), \\
    g_\omega(k) = g_\omega(\boldsymbol{\bar{z}}(k)) &\sim \mathcal{N}(\mu_\omega(k),\Sigma_\omega(k)).
\end{aligned}
\end{equation}

We show in Section~\ref{sec:uncertainty_propagation} how the mean and covariance values from Eqn.~\ref{eq:GP_approx} can be propagated to the remainder of the state-space using Sigma-Point Transforms, allowing us to fully realize our probabilistic motion model. 

\subsection{Weighted Ensemble GPR}
\label{sec:blend_gp}

Since the effects of \(\overline{\delta}_v\) and \(\overline{\delta}_\omega\) are terrain-dependent, GP hyperparameters trained under a single operating condition will lead to sub-optimal performance in a multi-terrain setting. Similar to~\cite{ensemble_gp}, we utilize the history of robot motions to compute a weighted sum of the distributions in Eqn.~\ref{eq:GP_approx}. Therefore, approximations of skid and slip at time instant \(k\) are produced for an ensemble of GPs trained over $M$ different terrains using:
\begin{align} \label{eq:weighted_gp_eqns}
[g_v(k),\,g_\omega(k)]^\intercal  
&\sim \mathcal{N}\left(\sum_{i=1}^Mw_i\boldsymbol{\mu}_i(k),\sum_{i=1}^Mw_i^2\boldsymbol{\Sigma}_i(k)\right).
\end{align}
Here, \(\boldsymbol{\mu}_i(k)\!\in\mathbb{R}^2\) and \(\boldsymbol{\Sigma}_i(k)\!\in\!\mathbb{R}^{2 \times 2}\) represent the output mean and covariance estimates of a GP for terrain \(i\), and \(w_i\!\in\mathbb{R}\) is the weight associated with that output. 

In order to compute the optimal set of weights \(\boldsymbol{w}^* = [w_1^*,\ldots, w_M^*]^\intercal\!\in\mathbb{R}^{M}\) at each iteration \(k\), we compare the predicted means with measured ground truth velocities from all GPs over a trajectory history of length \(K\). The resultant optimization problem has the following form:
\begin{align} \label{eq:convex_opt}
    \argmin_{\boldsymbol{w}} ||\boldsymbol{Y}_v&-\boldsymbol{F}_v\boldsymbol{w}||_2^2 + ||\boldsymbol{Y}_\omega-\boldsymbol{F}_\omega\boldsymbol{w}||_2^2 +  \alpha||\boldsymbol{w}-\boldsymbol{w}^{*}_{k-1}||_1 \nonumber \\
    &\textrm{s.t.} \quad 0\leq w_i \leq 1, \qquad \sum_{i=1}^{M}w_i=1,
\end{align}
where \(\boldsymbol{Y}_v\!\in\!\mathbb{R}^{K}\) and 
\(\boldsymbol{Y}_\omega\!\in\!\mathbb{R}^{K}\) represent ground truth velocities at each past time step, while \(\boldsymbol{F}_v\!\in\!\mathbb{R}^{K \times M}\) and \(\boldsymbol{F}_\omega\!\in\!\mathbb{R}^{K \times M}\) represent the corresponding GP mean predictions. The \(\mathcal{L}_1\) term in Eqn.~\ref{eq:convex_opt} penalizes deviations in the optimal weight estimates, \(\boldsymbol{w}^{*}\), from the previous solutions, \(\boldsymbol{w}^{*}_{k-1}\). Using these optimal weights, \(\boldsymbol{w}^{*}\), a convex combination of the outputs from all $M$ GPs best represents the terrain that the robot is currently navigating. 

\subsection{Uncertainty Propagation}

\label{sec:uncertainty_propagation}

The \textit{kinematic} sub state-space, \(\boldsymbol{q}\), is estimated based on Eqn.~\ref{eq:kinematic_continuous} from the robot velocities. Since these velocities are estimated probabilistically using GPR, the predicted future states also have a stochastic distribution~\cite{prop_uncertainty_2,prop_uncertainty_3}.

To efficiently compute the propagated state distribution, we make a simplifying assumption that the overall state-space of the robot is jointly Gaussian distributed,~\ie, \(\boldsymbol{x}\sim\mathcal{N}(\boldsymbol{\mu},\,\boldsymbol{\Sigma})\). Under such an assumption, Hewing et al.~\cite{car_gp_mpc} successfully demonstrated a miniature race car performing high-speed autonomous loops in a realistic simulator. In this work, we utilize Sigma-Point Transforms as a means to estimate the predictive distribution~\cite{ukf}.
Given the robot state dimension, \(n=5\), and a scalar tuning parameter, \(\lambda\), the mean and covariance of the predictive distribution is obtained via a weighted sum of Sigma-Point Transforms. As in~\cite{ostafew}, we chose \(4n+1\) Sigma Points with the corresponding weights:
\begin{equation}\label{eq:sigma_pt_weights}
    \mathbb{W}_0 = \frac{\lambda}{2n+\lambda},\quad
    \mathbb{W}_{1-4n+1} = \frac{0.5}{2n+\lambda}
\end{equation}


\begin{algorithm}[tb] \label{algo:UKF}
\setstretch{1.1} 
\caption{\textbf{Uncertainty State Propagation}}
\begin{algorithmic}[1]
\IRequire Prediction horizon, $N$
    \Statex \quad \, Control sequence, $\boldsymbol{u}=[\,\boldsymbol{u}(0),\ldots,\boldsymbol{u}(N-1)\,]^\intercal$
    \Statex \quad \, Initial state mean and covariance, $\boldsymbol{\mu}(0)$ and $\boldsymbol{\Sigma}(0)$
    \Statex \quad \, Scalar tuning parameter, $\lambda$
    \Statex \quad \, Sigma-Point weights, $\boldsymbol{\mathbb{W}} = [\mathbb{W}_0,\ldots,\mathbb{W}_{4n+1}]$
\Ensure{
    $\boldsymbol{\mu} = [\boldsymbol{\mu}(0),\ldots,\boldsymbol{\mu}(N)]^\intercal 
    ,\;\boldsymbol{\Sigma} = [\boldsymbol{\Sigma}(0),\ldots,\boldsymbol{\Sigma}(N)] $
    }
\For{$i = 0$ to $N-1$}
    \State $\boldsymbol{\mu}_{gp}(i) = [0, 0, 0, \mu_v{(i)}, \mu_w{(i)}]^\intercal$  
    \State $\boldsymbol{\Sigma}_{gp} = \text{diag}[0, 0, 0, \Sigma_v{(i)}, \Sigma_w{(i)}]$
    \State $\boldsymbol{a}(i) = [\boldsymbol{\mu}(i),\,\boldsymbol{\mu}_{gp}(i)]^T$
    \State $\boldsymbol{P}(i) = \text{diag}(\boldsymbol{\Sigma}(i),\,\boldsymbol{\Sigma}_{gp}(i))$
    \State $\boldsymbol{S}(i) = \sqrt{\boldsymbol{P}(i)}$
    \For{$j = 0$ to $4n$} \Comment{Sigma Points}
        \State $ \text{col}_s = j^{\text{th}}$ column of $\boldsymbol{S}(i)$
        \If{$j = 0$}
            \State $\boldsymbol{A}_j = \boldsymbol{a}(i)$ 
        \ElsIf{$j < 2n$}
            \State $\boldsymbol{A}_j = \boldsymbol{a}(i) + \sqrt{2n + \lambda} \; \text{col}_s$
        \Else
            \State $\boldsymbol{A}_j = \boldsymbol{a}(i) - \sqrt{2n + \lambda}  \; \text{col}_s$
        \EndIf
        \State $\boldsymbol{A}_j = (\boldsymbol{X}_j, \boldsymbol{M}_j)$
        \State $\boldsymbol{X}_{\text{next}, j} = \boldsymbol{f}(\boldsymbol{X}_j, \boldsymbol{u}(i)) + \boldsymbol{M}_j$
    \EndFor
    \State $\boldsymbol{\mu}(i+1) = \sum_{j=1}^{4n+1} 
    \mathbb{W}_j\,\boldsymbol{X}_{\text{next}, j}\;$ 
    \State $\boldsymbol{\Sigma}(i+1) = \sum_{j=1}^{4n+1} 
    \mathbb{W}_j([ \boldsymbol{X}_{\text{next}, j} - \boldsymbol{\mu}(i+1)] ^\intercal$ \\ \hspace{3.75cm} $\cdot \; [ \boldsymbol{X}_{\text{next}, j} - \boldsymbol{\mu}(i+1)])$
\EndFor
\end{algorithmic}
\vspace{-0.1em}
\end{algorithm}

In Algorithm $1$, we outline the steps necessary to apply this uncertainty propagation procedure for probabilistic motion modeling. 


\section{Experiments And Results}
\label{sec:results}
In this section, we compare the performance of the proposed GPR model against recent state-of-the-art kinematic motion models on a real-world dataset. A video showing the model's performance on an SSWMR navigating a lab floor and an artificial grass turf can also be found here: \url{https://www.youtube.com/watch?v=_rVy2aBp42c}.

\begin{figure*}[t]
    \vspace{2mm}
  \centering
    \subfigure[Asphalt]{
    \includegraphics[width=0.31\linewidth]{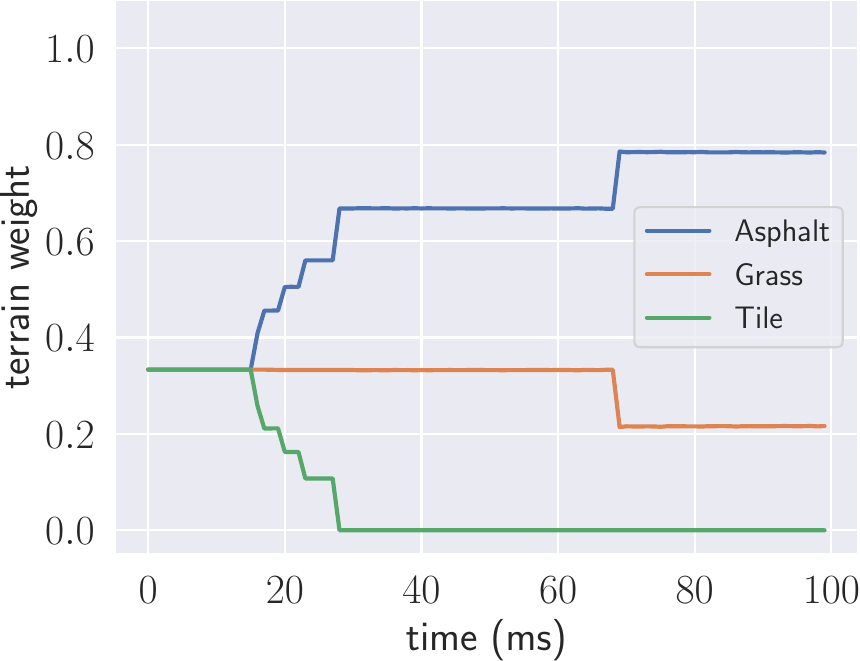}
    \label{fig:weight_asphalt}
    }
    \subfigure[Grass]{
    \includegraphics[width=0.31\linewidth]{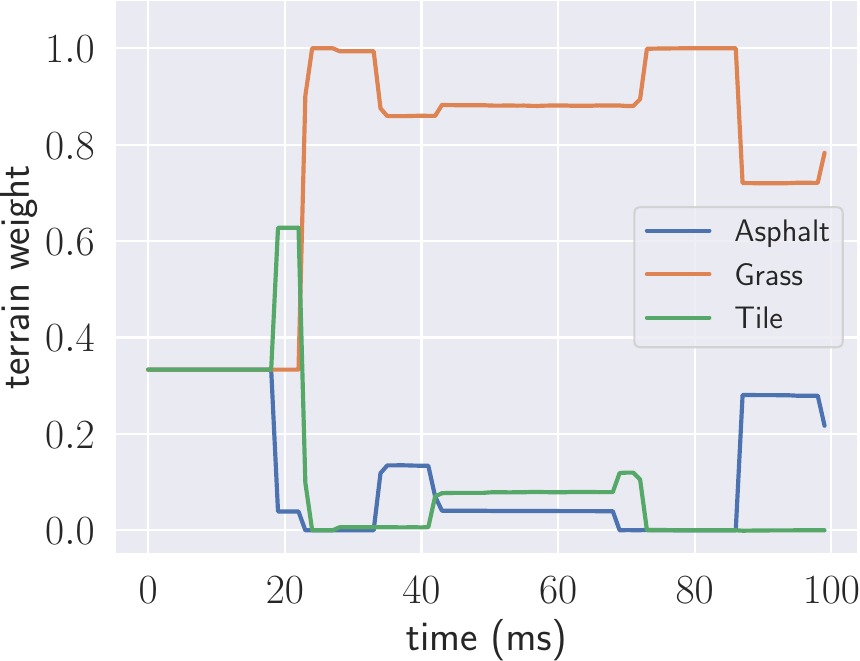}
    \label{fig:weight_grass}
    }
    \subfigure[Tile]{
    \includegraphics[width=0.31\linewidth]
    {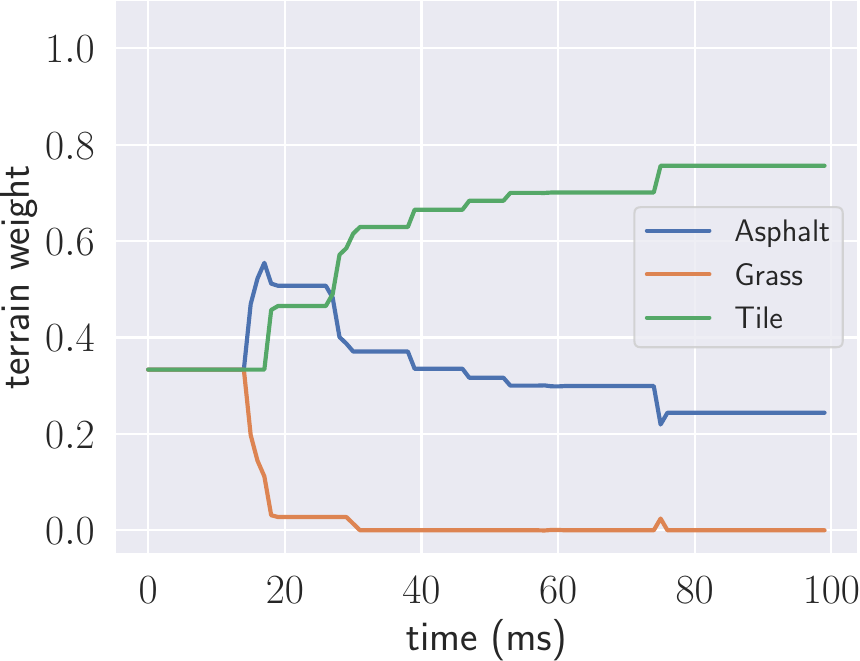}
    \label{fig:weight_tile}
    }
  \caption{Illustration of the optimal weight estimates for each terrain condition using ensemble Gaussian Process regression.}
  \label{fig:ensemble_gp}
  \vspace{-0.7em}
\end{figure*}




\subsection{Benchmarking Dataset Description} \label{sec:dataset}

\begin{figure}[t]
    \centering\includegraphics[width=0.9\columnwidth,height=0.5\columnwidth]
    {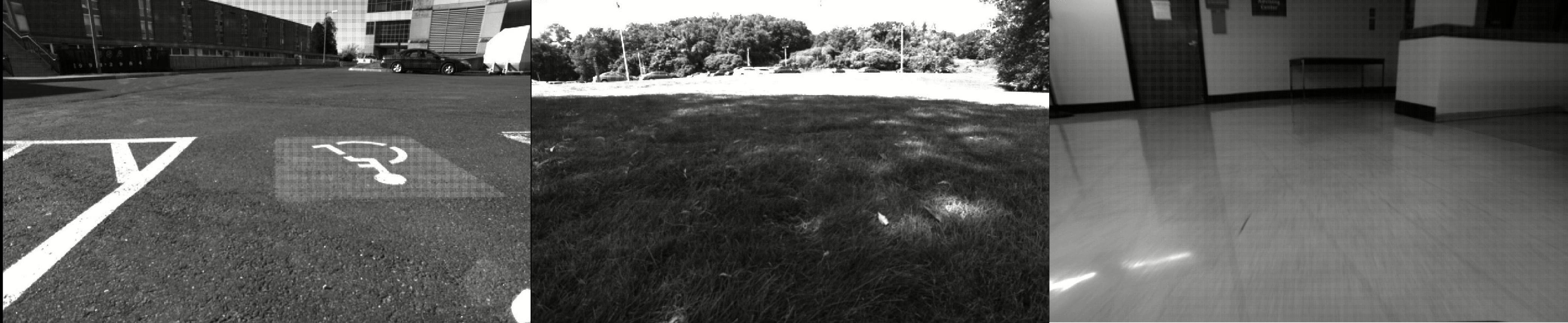}
        \caption{Terrain snapshots: Asphalt, Grass, and Tile~\cite{rabiee}.}

    \label{fig:terrain_snapshots}
\end{figure}

As shown in Fig.~\ref{fig:terrain_snapshots},  Rabiee et al.~\cite{rabiee} released a benchmark dataset of SSWMR motion on three different terrains: Tile, Asphalt, and Grass for the SSWMR Clearpath Jackal.

The data comprised of commanded velocities and ground truth robot positions generated via ORB-SLAM2~\cite{orb_slam}. We generated ground truth robot velocities by passing the ORB-SLAM2 trajectories through a first-order low-pass-filter and subsequently applying finite backward differences. As discussed in the sections above, commanded and ground truth velocities form the basis of the GPR used in our motion model.

\subsection{Model Training Process}
\label{sec:model_training_process} 
All the code for our proposed motion modeling approach was implemented in Python. Parameters, \([c_1,\ldots,c_6]^\intercal\) were estimated as in Eqn.~\ref{eq:c1_to_c6} using the SciPy~\cite{scipy} library. GP model training and inference was performed using GPyTorch~\cite{gpytorch}, since it allows for fast variance predictions and kernel operations~\cite{love_gp,keops}. Based on the GP input features, we find 500 unique cluster centers using a Gaussian Mixture Model (GMM). Data points with the shortest Euclidean distance to each of the cluster centers are chosen as the training points for the optimal GP hyperparameter estimation process~\cite{safe_control_gym}.

\subsection{Model 
Generalization}\label{sec:model_generalization}

As shown in Section~\ref{sec:blend_gp}, by combining the outputs of different GP models in a weighted sum fashion, we are able to apply our motion model to diverse and unseen terrain conditions. We solve Eqn.~\ref{eq:convex_opt} with the configuration values of $K=10$, $M=3$, and a discretization of $0.1$ seconds between time-steps. The resultant optimization problem was solved using the OSQP solver~\cite{osqp} with the CVXPY~\cite{cvxpy} interface. The ensemble GP results are provided in Fig.~\ref{fig:ensemble_gp}. 

When the robot begins its motion at time $t=0$, we warmstart the optimization problem by assigning uniform probability to all three terrains. As the robot motion history builds up, we see the optimizer converge to the correct terrain in less than $100$\,ms. It should be noted that we never converge to the truly ideal solution of a weight of $1$ for the terrain of traversal and $0$ for the rest. We attribute this to the inherently noisy nature of a real-world dataset.

\begin{table}[t]
    \caption{Mean \% error in predicted position}
    \label{tab:vel_perc_err}
    \centering
    \footnotesize
    \begin{tabular}{l|ccc|ccc}
        \toprule
        & \multicolumn{3}{c}{Angular} & \multicolumn{3}{c}{Linear} \\
        \cmidrule(lr){1-7} 
         & Asphalt & Grass & Tile & Asphalt & Grass & Tile \\
        \midrule
        EDD5 & $17.6$ & $18.9$ & $21.1$ & $14.2$ & $6.2$ & $11.6$ \\
        GP & $\textbf{5.7}$ & $\textbf{5.6}$ & $\textbf{10.9}$ & $\textbf{5.8}$ & $\textbf{5.7}$ & $\textbf{5.0}$ \\
        \bottomrule
    \end{tabular}
    \vspace{-3.3mm}
\end{table}

\begin{table}[t]
    \caption{Mean Absolute Errors in predicted Velocity}
    \label{tab:vel_abs_err}
    \centering
    \footnotesize
    \begin{tabular}{l|ccc|ccc}
        \toprule
        & \multicolumn{3}{c}{Angular (rad/s)} & \multicolumn{3}{c}{Linear (m/s)} \\
        \cmidrule(lr){1-7} 
         & Asphalt & Grass & Tile & Asphalt & Grass & Tile \\
        \midrule
        EDD5 & $0.19$ & $0.16$ & $0.30$ & $0.23$ & $\textbf{0.07}$ & $0.22$ \\
        GP & $\textbf{0.02}$ & $\textbf{0.03}$ & $\textbf{0.06}$ & $\textbf{0.12}$ & $0.09$ & $\textbf{0.13}$ \\
        \bottomrule
    \end{tabular}
    \vspace{-1.7mm}
\end{table}

\subsection{Benchmarking Model 
Accuracy}\label{sec:benchmarking_model_accuracy}



 Baril et al.~\cite{baril} described five popular parameterizations of the Jacobian matrix in Eqn.~\ref{eq:kinematic_models}, thereby yielding five kinematic models: the ideal differential drive (IDD), the extended differential drive with two (EDD2) and five parameters (EDD5), the radius of curvature (ROC), and the fully linear (FL) model. In this section, we compare the prediction accuracy of our GP approach against these kinematic models and report only on the best performing model (EDD5) in Tables~\ref{tab:vel_perc_err} \&~\ref{tab:vel_abs_err}.

In order to generate the mean error results shown in Tables~\ref{tab:vel_perc_err} \&~\ref{tab:vel_abs_err}, we elect a $1$-second prediction horizon and discretize it into $N$ time-steps. We predict the state of the robot, $\boldsymbol{x}(k+N)_{pred}$, for a sequence of commands, $\boldsymbol{u}=[\boldsymbol{u}(0),\ldots,\boldsymbol{u}({N-1})]^\intercal$, using both the GP-based and EED5 motion models. Given the ground truth robot state, $\boldsymbol{x}(k+N)_{gt}$, estimated via ORB-SLAM2, the prediction error is defined as:
\begin{equation}
    \nonumber
    \boldsymbol{e}(k+N) =  \boldsymbol{x}(k+N)_{pred} - \boldsymbol{x}(k+N)_{gt}.
\end{equation}

For each terrain, we compute errors over thousands of $1$-second trajectories obtained by performing a moving horizon sweep across their respective datasets. Furthermore, in Table~\ref{tab:vel_perc_err}, the positional errors are normalized based on the displacement at the end of the horizon. 

We observe that the GP-based motion model performs significantly better than the EDD5 model in terms of both rotational and translational errors. Kinematic models being inherently linear are unable to accurately capture the non-linear effects of tire slip and skid. The GP dynamic model in contrast, reasons over wheel-ground contact interactions explicitly by learning the effects of unmodeled tire forces, $\overline{\delta}_v$ and $\overline{\delta}_\omega$. The dynamic effects of robot accelerations are thus better captured by the proposed model leading to improvements in predictive performance.

\subsection{Probabilistic State Estimation}\label{sec:probabilistic_stae_estimation}

\begin{figure}[t]
    \vspace{2mm}
    \centering
    \subfigure[Sigma-Point Transform]{
    \includegraphics[width=0.7\columnwidth]{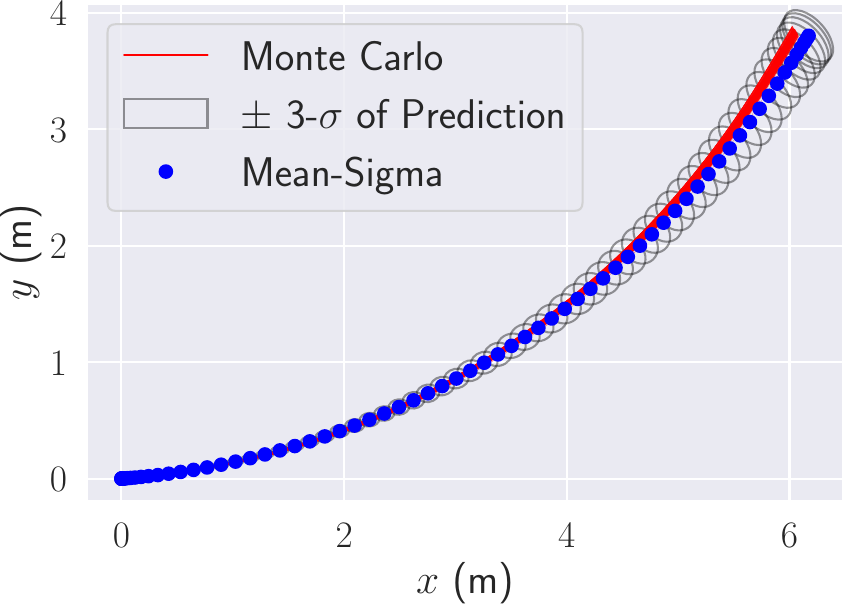}
    \label{fig:uprop:2}
    }
    \subfigure[Taylor Series Expansion]{
    \includegraphics[width=0.7\columnwidth]{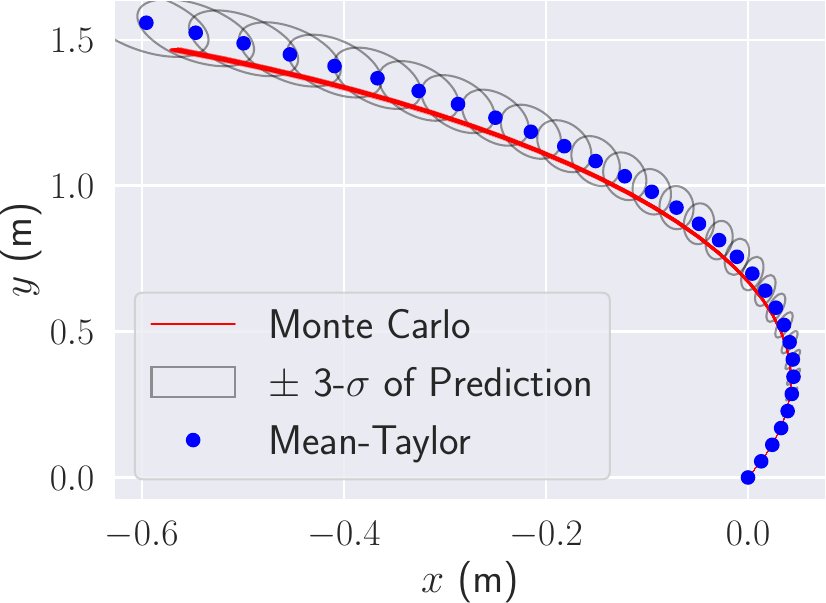}
    \label{fig:uprop:3}
    }
    \caption{Uncertainty state propagation for two robot trajectories via linear (bottom) and non-linear (top) methods.}
    \label{fig:uprop}
    \vspace{-2.3mm}
\end{figure}

As shown in Section~\ref{sec:method}, we predict the effects of tire skid and slip on the robot velocities using GPR. Under the assumption of the robot state-space $\boldsymbol{x}$ being jointly Gaussian, we approximated the distribution of robot kinematics $\boldsymbol{q}$ from GP variance in Algorithm $1$. In this section, we validate the accuracy of that assumption.

Fig.~\ref{fig:uprop} shows the evolution of the distribution of
robot positions  for two random trajectories using non-linear (Sigma-Point Transforms) and linear (Taylor Series Expansion) methods. At each time step, predictions of next state velocities are made using the GP-based motion model. The resultant Gaussian distributed robot positions are obtained via Algorithm $1$ in Section~\ref{sec:uncertainty_propagation}. The true distribution is approximated via $250$ Monte-Carlo samples, indicated in red. As the prediction horizon becomes larger, we observe an increase in the uncertainty surrounding position estimates, as evidenced by the $3\!\!-\!\!\sigma$ ellipse size. Despite this uncertainty, the Monte-Carlo trajectories are completely contained within the ellipses. In turn, we conclude that for relatively small prediction horizons, the Gaussian distribution obtained from the Sigma-Point Transforms closely approximates the true distribution of the robot's position trajectory.

\subsection{Evaluating Tire Slip Estimation}\label{estimating_terrain_based_nonlinearities}


\begin{figure}[t]
    \vspace{2mm}
    \centering
    \subfigure[GP-based Model]{
        \includegraphics[width=0.78\columnwidth]{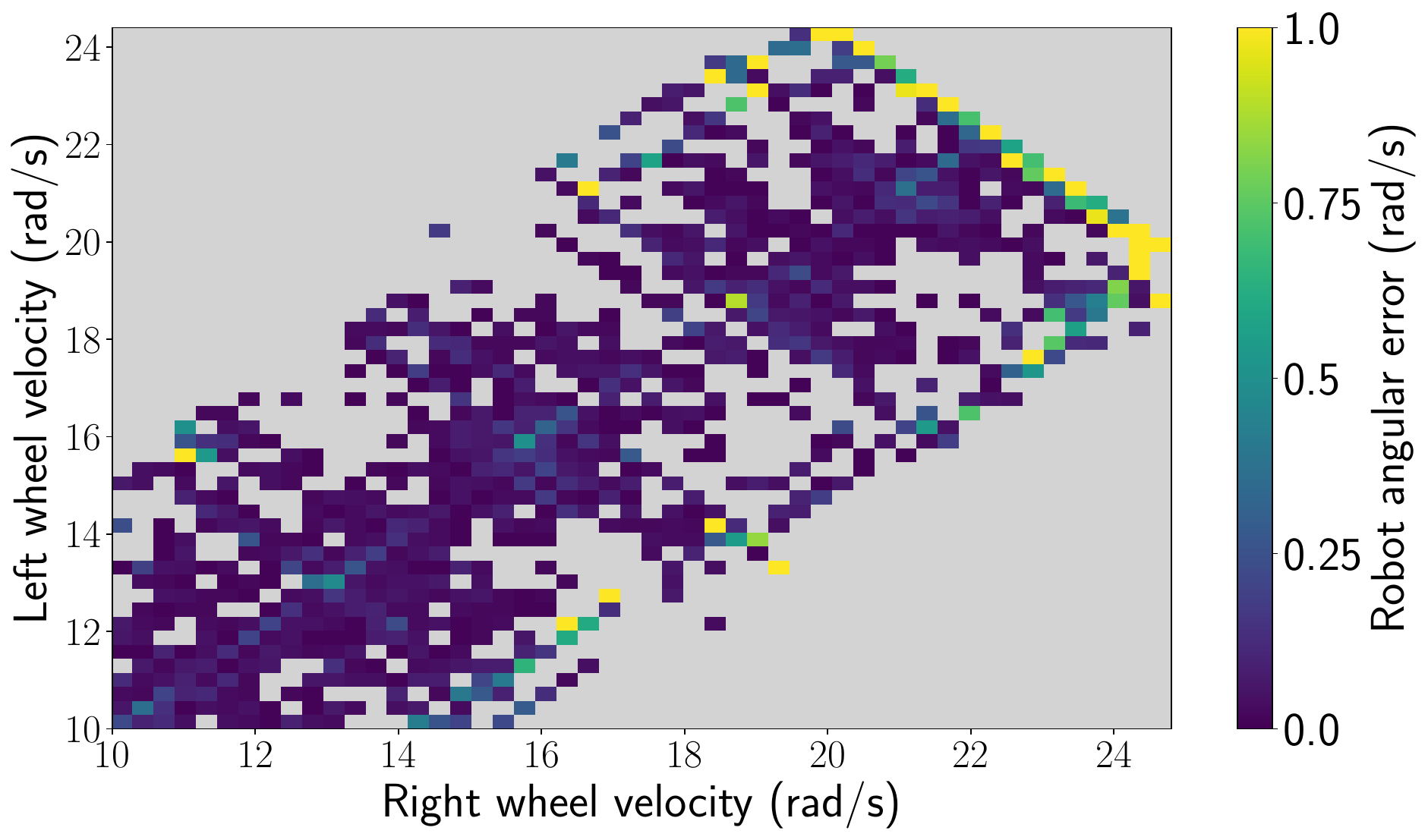}
        \label{fig:heat_map:gp}
    }
    \subfigure[Kinematic Model]{
        \includegraphics[width=0.78\columnwidth]{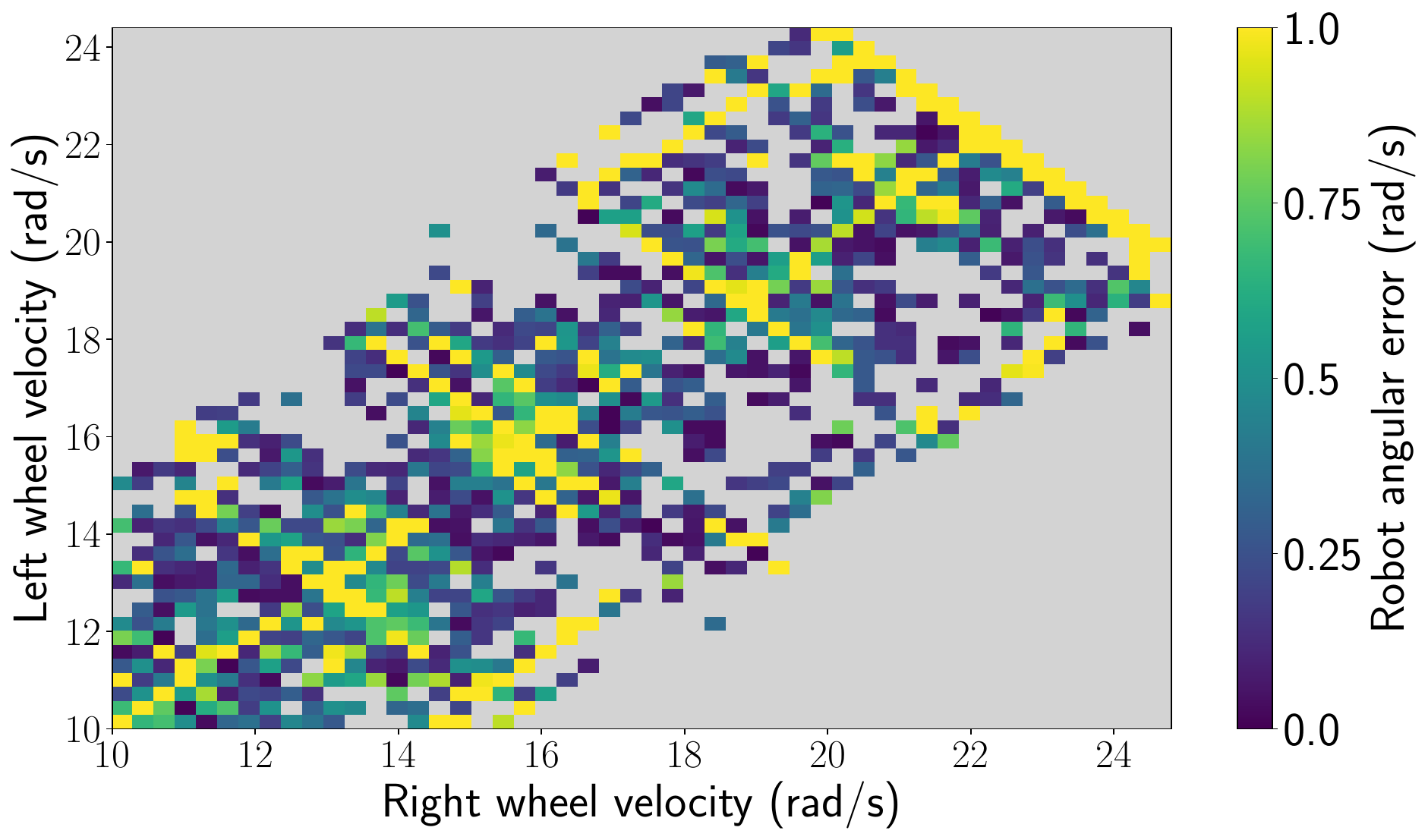}
        \label{fig:heat_map:kin}
    }
    \caption{Prediction error based on commanded velocities.}
    \label{fig:heat_map}
    \vspace{-3.5mm}
\end{figure}

In Eqn.~\ref{eq:slip_definition}, the tire slip ratio and slip angle are described as a nonlinear function of wheel velocities. Baril et al.~\cite{baril} analyzed motion prediction errors for different kinematic models. Their key observation was that when turning at high linear velocities, kinematic motion models see a spike in angular motion prediction errors. Brach et al.~\cite{slip_is_non_linear} attributed this effect to increased tire slip at high operational velocities. We investigate this in Fig.~\ref{fig:heat_map} for the Asphalt terrain. It must be noted that similar trends were observed for the Grass and Tile terrain.

Our findings demonstrate that the GP-based model is generally more accurate in its prediction of angular velocity profiles. The dark yellow patches in Fig.~\ref{fig:heat_map:kin} are more evident for the kinematic model, suggesting the difference between commanded wheel velocities is higher. Side-slip is a measure of the misalignment between vehicle orientation and trajectory. As shown in~\cite{side_slip}, extreme turning motions are prone to higher magnitudes of side-slip. In contrast to the GP-based model, the kinematic models in~\cite{baril} were unable to accurately capture these effects leading to higher angular prediction error. This reinforces our choice of using GPR as an effective tool for robot system identification. 

\section{Conclusion}
\label{sec:conclusion}
In this paper, we introduced a probabilistic motion model for SSWMRs to estimate the effects of tire-slip and skid on predicted robot poses. Experimental results on a real-world dataset provided insight into our model's significant improvements in performance over conventional kinematic motion models for both linear and angular velocity prediction. Additionally, we presented an approach involving an ensemble of GPs to generalize our motion model's applicability to previously unseen terrain conditions. 

A limitation of the presented work is that the computational complexity of the GPR inference step scales with the number of terrains. To mitigate this, we plan to explore additional acceleration methods for GP kernel operations on a GPU. Furthermore, we intend to incorporate our probabilistic motion model into the sampling-based Stochastic MPC framework introduced in our prior work~\cite{shameless_plug}.

\bibliographystyle{IEEEtran}
\bibliography{IEEEabrv,references}

\begin{thebibliography}{10}
\providecommand{\url}[1]{#1}
\csname url@rmstyle\endcsname
\providecommand{\newblock}{\relax}
\providecommand{\bibinfo}[2]{#2}
\providecommand\BIBentrySTDinterwordspacing{\spaceskip=0pt\relax}
\providecommand\BIBentryALTinterwordstretchfactor{4}
\providecommand\BIBentryALTinterwordspacing{\spaceskip=\fontdimen2\font plus
\BIBentryALTinterwordstretchfactor\fontdimen3\font minus
  \fontdimen4\font\relax}
\providecommand\BIBforeignlanguage[2]{{%
\expandafter\ifx\csname l@#1\endcsname\relax
\typeout{** WARNING: IEEEtran.bst: No hyphenation pattern has been}%
\typeout{** loaded for the language `#1'. Using the pattern for}%
\typeout{** the default language instead.}%
\else
\language=\csname l@#1\endcsname
\fi
#2}}

\bibitem{SSMR_Survey}
R.~Khan, F.~M. Malik, A.~Raza, and N.~Mazhar, ``Comprehensive study of
  skid-steer wheeled mobile robots: Development and challenges,''
  \emph{Industrial Robot: the international journal of robotics research and
  application}, vol.~48, no.~1, pp. 142--156, 2021.

\bibitem{vehicle_dynamics}
R.~Rajamani, \emph{Vehicle dynamics and control}.\hskip 1em plus 0.5em minus
  0.4em\relax Springer Science \& Business Media, 2011.

\bibitem{rasmussen}
C.~K. Williams and C.~E. Rasmussen, \emph{Gaussian processes for machine
  learning}.\hskip 1em plus 0.5em minus 0.4em\relax MIT press Cambridge, MA,
  2006, vol.~2, no.~3.

\bibitem{ostafew}
C.~J. Ostafew, A.~P. Schoellig, and T.~D. Barfoot, ``Robust constrained
  learning-based nmpc enabling reliable mobile robot path tracking,'' \emph{The
  International Journal of Robotics Research}, vol.~35, no.~13, pp. 1547--1563,
  2016.

\bibitem{eth_gpmpc}
L.~Hewing, J.~Kabzan, and M.~N. Zeilinger, ``Cautious model predictive control
  using gaussian process regression,'' \emph{IEEE Transactions on Control
  Systems Technology}, vol.~28, no.~6, pp. 2736--2743, 2019.

\bibitem{ukf}
S.~J. Julier and J.~K. Uhlmann, ``Unscented filtering and nonlinear
  estimation,'' \emph{Proceedings of the IEEE}, vol.~92, no.~3, pp. 401--422,
  2004.

\bibitem{slip_is_non_linear}
R.~Brach and M.~Brach, ``The tire-force ellipse (friction ellipse) and tire
  characteristics,'' SAE Technical Paper, Tech. Rep., 2011.

\bibitem{orb_slam}
R.~Mur-Artal and J.~D. Tard{\'o}s, ``Orb-slam2: An open-source slam system for
  monocular, stereo, and rgb-d cameras,'' \emph{IEEE transactions on robotics},
  vol.~33, no.~5, pp. 1255--1262, 2017.

\bibitem{taskin_ssmr}
T.~J. Carlone, J.~J. Anderson, J.~L. Amato, V.~D. Dimitrov, and T.~Padir,
  ``Kinematic control of a planetary exploration rover over rough terrain,'' in
  \emph{2013 IEEE International Conference on Systems, Man, and
  Cybernetics}.\hskip 1em plus 0.5em minus 0.4em\relax IEEE, 2013, pp.
  4488--4493.

\bibitem{low_and_wang_kinematic}
D.~Wang and C.~B. Low, ``Modeling and analysis of skidding and slipping in
  wheeled mobile robots: Control design perspective,'' \emph{IEEE Transactions
  on Robotics}, vol.~24, no.~3, pp. 676--687, 2008.

\bibitem{martinez_kinematic}
E.~Mart{\'\i}nez-Garc{\'\i}a and R.~Torres-C{\'o}rdoba, ``4wd skid-steer
  trajectory control of a rover with spring-based suspension analysis: direct
  and inverse kinematic parameters solution,'' in \emph{Intelligent Robotics
  and Applications: Third International Conference, ICIRA 2010, Shanghai,
  China, November 10-12, 2010. Proceedings, Part I 3}.\hskip 1em plus 0.5em
  minus 0.4em\relax Springer, 2010, pp. 453--464.

\bibitem{baril}
D.~Baril, V.~Grondin, S.-P. Desch{\^e}nes, J.~Laconte, M.~Vaidis, V.~Kubelka,
  A.~Gallant, P.~Giguere, and F.~Pomerleau, ``Evaluation of skid-steering
  kinematic models for subarctic environments,'' in \emph{2020 17th Conference
  on Computer and Robot Vision (CRV)}.\hskip 1em plus 0.5em minus 0.4em\relax
  IEEE, 2020, pp. 198--205.

\bibitem{yu_dynamic}
W.~Yu, O.~Chuy, E.~G. Collins, and P.~Hollis, ``Dynamic modeling of a
  skid-steered wheeled vehicle with experimental verification,'' in \emph{2009
  IEEE/RSJ International Conference on Intelligent Robots and Systems}.\hskip
  1em plus 0.5em minus 0.4em\relax IEEE, 2009, pp. 4212--4219.

\bibitem{kozlowski_sim_dynamic}
K.~Koz{\l}owski and D.~Pazderski, ``Modeling and control of a 4-wheel
  skid-steering mobile robot,'' \emph{International journal of applied
  mathematics and computer science}, vol.~14, no.~4, pp. 477--496, 2004.

\bibitem{seegmiller_dynamic}
N.~Seegmiller and A.~Kelly, ``Modular dynamic simulation of wheeled mobile
  robots,'' in \emph{Field and Service Robotics: Results of the 9th
  International Conference}.\hskip 1em plus 0.5em minus 0.4em\relax Springer,
  2015, pp. 75--89.

\bibitem{rabiee}
S.~Rabiee and J.~Biswas, ``A friction-based kinematic model for skid-steer
  wheeled mobile robots,'' in \emph{2019 International Conference on Robotics
  and Automation (ICRA)}.\hskip 1em plus 0.5em minus 0.4em\relax IEEE, 2019,
  pp. 8563--8569.

\bibitem{gpytorch}
J.~Gardner, G.~Pleiss, K.~Q. Weinberger, D.~Bindel, and A.~G. Wilson,
  ``Gpytorch: Blackbox matrix-matrix gaussian process inference with gpu
  acceleration,'' \emph{Advances in neural information processing systems},
  vol.~31, 2018.

\bibitem{huskic}
G.~Huski{\'c}, S.~Buck, M.~Herrb, S.~Lacroix, and A.~Zell, ``High-speed path
  following control of skid-steered vehicles,'' \emph{The International Journal
  of Robotics Research}, vol.~38, no.~9, pp. 1124--1148, 2019.

\bibitem{jfr_gp_sswmr}
J.~Wang, M.~T. Fader, and J.~A. Marshall, ``Learning-based model predictive
  control for improved mobile robot path following using gaussian processes and
  feedback linearization,'' \emph{Journal of Field Robotics}, 2023.

\bibitem{teb}
C.~R{\"o}smann, W.~Feiten, T.~W{\"o}sch, F.~Hoffmann, and T.~Bertram,
  ``Trajectory modification considering dynamic constraints of autonomous
  robots,'' in \emph{ROBOTIK 2012; 7th German Conference on Robotics}.\hskip
  1em plus 0.5em minus 0.4em\relax VDE, 2012, pp. 1--6.

\bibitem{Everett_SSMR}
X.~Cai, M.~Everett, L.~Sharma, P.~R. Osteen, and J.~P. How, ``Probabilistic
  traversability model for risk-aware motion planning in off-road
  environments,'' in \emph{2023 IEEE/RSJ International Conference on
  Intelligent Robots and Systems (IROS)}.\hskip 1em plus 0.5em minus
  0.4em\relax IEEE, 2023, pp. 11\,297--11\,304.

\bibitem{ensemble_gp}
T.~Nagy, A.~Amine, T.~X. Nghiem, U.~Rosolia, Z.~Zang, and R.~Mangharam,
  ``Ensemble gaussian processes for adaptive autonomous driving on
  multi-friction surfaces,'' \emph{arXiv preprint arXiv:2303.13694}, 2023.

\bibitem{dynamic_unicycle}
C.~De~La~Cruz and R.~Carelli, ``Dynamic modeling and centralized formation
  control of mobile robots,'' in \emph{IECON 2006-32nd annual conference on
  IEEE industrial electronics}.\hskip 1em plus 0.5em minus 0.4em\relax IEEE,
  2006, pp. 3880--3885.

\bibitem{ssmr_modeling_3d}
{\'A}.~J. Prado, M.~Torres-Torriti, J.~Yuz, and F.~A. Cheein, ``Tube-based
  nonlinear model predictive control for autonomous skid-steer mobile robots
  with tire--terrain interactions,'' \emph{Control Engineering Practice}, vol.
  101, p. 104451, 2020.

\bibitem{slotine}
J.-J.~E. Slotine, W.~Li, \emph{et~al.}, \emph{Applied nonlinear control}.\hskip
  1em plus 0.5em minus 0.4em\relax Prentice hall Englewood Cliffs, NJ, 1991,
  vol. 199, no.~1.

\bibitem{least_squares_params}
F.~Reyes and R.~Kelly, ``On parameter identification of robot manipulators,''
  in \emph{Proceedings of International Conference on Robotics and Automation},
  vol.~3.\hskip 1em plus 0.5em minus 0.4em\relax IEEE, 1997, pp. 1910--1915.

\bibitem{car_gp_mpc}
L.~Hewing, A.~Liniger, and M.~N. Zeilinger, ``Cautious nmpc with gaussian
  process dynamics for autonomous miniature race cars,'' in \emph{2018 European
  Control Conference (ECC)}.\hskip 1em plus 0.5em minus 0.4em\relax IEEE, 2018,
  pp. 1341--1348.

\bibitem{acados}
R.~Verschueren, G.~Frison, D.~Kouzoupis, J.~Frey, N.~v. Duijkeren, A.~Zanelli,
  B.~Novoselnik, T.~Albin, R.~Quirynen, and M.~Diehl, ``acados—a modular
  open-source framework for fast embedded optimal control,'' \emph{Mathematical
  Programming Computation}, vol.~14, no.~1, pp. 147--183, 2022.

\bibitem{forcespro}
A.~Zanelli, A.~Domahidi, J.~Jerez, and M.~Morari, ``Forces nlp: an efficient
  implementation of interior-point methods for multistage nonlinear nonconvex
  programs,'' \emph{International Journal of Control}, vol.~93, no.~1, pp.
  13--29, 2020.

\bibitem{prop_uncertainty_2}
M.~Kuss, ``Gaussian process models for robust regression, classification, and
  reinforcement learning,'' Ph.D. dissertation, Technische Universit{\"a}t
  Darmstadt, 2006.

\bibitem{prop_uncertainty_3}
M.~P. Deisenroth, \emph{Efficient reinforcement learning using Gaussian
  processes}.\hskip 1em plus 0.5em minus 0.4em\relax KIT Scientific Publishing,
  2010, vol.~9.

\bibitem{scipy}
P.~Virtanen, R.~Gommers, T.~E. Oliphant, M.~Haberland, T.~Reddy, D.~Cournapeau,
  E.~Burovski, P.~Peterson, W.~Weckesser, J.~Bright, \emph{et~al.}, ``Scipy
  1.0: fundamental algorithms for scientific computing in python,''
  \emph{Nature methods}, vol.~17, no.~3, pp. 261--272, 2020.

\bibitem{love_gp}
G.~Pleiss, J.~Gardner, K.~Weinberger, and A.~G. Wilson, ``Constant-time
  predictive distributions for gaussian processes,'' in \emph{International
  Conference on Machine Learning}.\hskip 1em plus 0.5em minus 0.4em\relax PMLR,
  2018, pp. 4114--4123.

\bibitem{keops}
J.~Ragan-Kelley, A.~Adams, D.~Sharlet, C.~Barnes, S.~Paris, M.~Levoy,
  S.~Amarasinghe, and F.~Durand, ``Halide: Decoupling algorithms from schedules
  for high-performance image processing,'' \emph{Communications of the ACM},
  vol.~61, no.~1, pp. 106--115, 2017.

\bibitem{safe_control_gym}
Z.~Yuan, A.~W. Hall, S.~Zhou, L.~Brunke, M.~Greeff, J.~Panerati, and A.~P.
  Schoellig, ``Safe-control-gym: A unified benchmark suite for safe
  learning-based control and reinforcement learning,'' \emph{arXiv preprint
  arXiv:2109.06325}, 2021.

\bibitem{osqp}
B.~Stellato, G.~Banjac, P.~Goulart, A.~Bemporad, and S.~Boyd, ``Osqp: An
  operator splitting solver for quadratic programs,'' \emph{Mathematical
  Programming Computation}, vol.~12, no.~4, pp. 637--672, 2020.

\bibitem{cvxpy}
S.~Diamond and S.~Boyd, ``Cvxpy: A python-embedded modeling language for convex
  optimization,'' \emph{The Journal of Machine Learning Research}, vol.~17,
  no.~1, pp. 2909--2913, 2016.

\bibitem{side_slip}
D.~Chindamo, B.~Lenzo, and M.~Gadola, ``On the vehicle sideslip angle
  estimation: a literature review of methods, models, and innovations,''
  \emph{applied sciences}, vol.~8, no.~3, p. 355, 2018.

\bibitem{shameless_plug}
A.~Trivedi, S.~Bazzi, M.~Zolotas, and T.~Pad{\i}r, ``Probabilistic dynamic
  modeling and control for skid-steered mobile robots in off-road
  environments,'' in \emph{2023 IEEE International Conference on Assured
  Autonomy (ICAA)}.\hskip 1em plus 0.5em minus 0.4em\relax IEEE, 2023, pp.
  57--60.

\end{thebibliography}

\end{document}